\documentclass{article}

\usepackage[final]{nips_2017}

\usepackage[utf8]{inputenc} % allow utf-8 input
\usepackage[T1]{fontenc}    % use 8-bit T1 fonts
\usepackage{hyperref}       % hyperlinks
\usepackage{url}            % simple URL typesetting
\usepackage{booktabs}       % professional-quality tables
\usepackage{amsfonts}       % blackboard math symbols
\usepackage{nicefrac}       % compact symbols for 1/2, etc.
\usepackage{microtype}      % microtypography

\usepackage{caption}
\usepackage{subcaption}
\usepackage{bm}
\usepackage{dsfont}
\usepackage{amsmath}
\usepackage{eqparbox}
\usepackage{float}
\usepackage{algorithm}
\usepackage{graphicx}
\usepackage[noend]{algorithmic}
\usepackage{appendix}

\title{Comparing Generative Adversarial Network Techniques for Image Creation and Modification}

\author{
  Mathijs Pieters \\
  Institute of Artificial Intelligence\\
  University of Groningen, The Netherlands\\
  \texttt{m.t.pieters@student.rug.nl}
  \And
  Marco Wiering \\
  Institute of Artificial Intelligence\\
  University of Groningen, The Netherlands\\
  \texttt{m.a.wiering@rug.nl}
}
 
\begin{document}

\maketitle

\begin{abstract}
{Generative adversarial networks (GANs) have demonstrated to be successful at generating realistic real-world images. In this paper we compare various GAN techniques, both supervised and unsupervised. The effects on training stability of different objective functions are compared. We add an encoder to the network, making it possible to encode images to the latent space of the GAN. The generator, discriminator and encoder are parameterized by deep convolutional neural networks. For the discriminator network we experimented with using the novel Capsule Network, a state-of-the-art technique for detecting global features in images. Experiments are performed using a digit and face dataset, with various visualizations illustrating the results. The results show that using the encoder network it is possible to reconstruct images. With the conditional GAN we can alter visual attributes of generated or encoded images. The experiments with the Capsule Network as discriminator result in generated images of a lower quality, compared to a standard convolutional neural network.}
\end{abstract}

\section{Introduction}\label{sec:introduction}
Generative Adversarial Networks (GANs) \citep{generative} are a subclass of generative models that have received a lot of attention because of their ability to generate realistic high quality images. The GAN setup consists of two networks, a generator and discriminator, that compete against each other in a two-player minimax game. An analogy for this minimax game using the production of money is as follows. The generator is a counterfeiter that aims to create realistic money, whereas the discriminator's aim is to differentiate money created by the generator from real money. Both systems are trained simultaneously, and the competition should improve the systems until the generator produces counterfeits that are indistinguishable from real money. The generator does not have access to the real data, it only learns from the feedback from the discriminator. 

%TODO: relate conditional GAN to money analogy
 More recently, people started to apply the GAN framework to other problems, such as text generation and image-to-image translation \citep{zhang2017adversarial, image_to_image}. Since the introduction of the GAN, many variants have been proposed. A lot of research has been devoted to finding GAN algorithms that are more stable \citep{wasserstein, improved_wasserstein, berthelot2017began}. Although many papers claim to have found a significant improvement, \cite{equal_gans} show that there is no evidence to support these claims.
 
 %TODO: improve explanation of different GANs
 In this research we provide an overview of various GAN techniques. We compare standard GANs \citep{generative} with conditional GANs \citep{conditional_gan}, and supervised networks with unsupervised networks \citep{infogan}. Furthermore, we use an encoder making it possible to generate images that resemble specific images. This approach is very similar to the encoder network in a variational autoencoder \citep{kingma2013auto}. Finally, we compare a deep convolutional neural network with the novel Capsule Network as parameterization of the discriminator. These comparisons are performed on two datasets. The first dataset is MNIST \citep{mnist}, a dataset with images of handwritten digits. The second dataset we used is CelebA \citep{celebA}, containing images of faces. 
 
We explain the used methods in section \ref{sec:method}, followed by section \ref{sec:experiments} where we explain the experimental setup and show the results. Finally, in section \ref{sec:conclusion} we draw a conclusion and propose future research.
\section{Methods}\label{sec:method}

\subsection{Generative Adversarial Networks}\label{sec:GAN}
\begin{figure*}
	\centering
	\includegraphics[width=.8\linewidth]{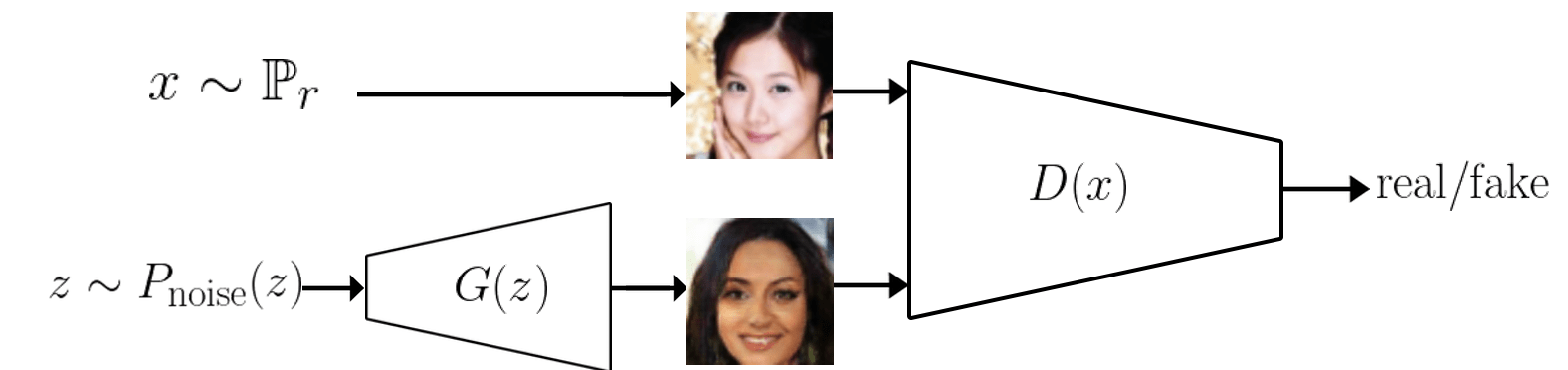}
	\caption{Graphical representation of a GAN.}
	\label{img:gan}
\end{figure*}

A Generative Adversarial Network consists of two competing networks, the \textit{generator} and the \textit{discriminator}. The generator tries to learn a mapping from a noise distribution to the real data distribution $\mathds{P}_r$, while the discriminator's task is to distinguish real data samples from samples generated by the generator $\mathds{P}_g$. The flow of data in a GAN is illustrated in Figure \ref{img:gan}. Formally, the generator $G$ transforms a noise sample $\bm{z}$ into a sample $\bm{\hat{x}} = G(\bm{z})$ (where $\bm{z} \sim P_{\text{noise}}(\bm{z})$ is sampled from a certain noise distribution such as a uniform distribution). This noise sample $\bm{z}$ is also referred to as the latent vector. The discriminator, indicated by $D(\bm{x})$, learns the probability that $\bm{x}$ originates from the real data distribution, rather than from $\mathds{P}_g$. $D$ is trained such that it maximizes the probability of classifying the samples from $\mathds{P}_r$ as real, and the samples from $\mathds{P}_g$ as fake. At the same time, $G$ is trained to minimize $\log(1- D(\bm{\hat{x}}))$. The lower this value, the higher $D(\bm{\hat{x}})$, indicating a high probability that $\bm{\hat{x}}$ is considered to be generated by the real data distribution.
 The training of the generator $G$ and the discriminator $D$ is defined as the following minimax game: \begin{equation} \min_{G} \max_{D} \mathds{E}_{\bm{x} \sim \mathds{P}_r} \lbrack \log D(\bm{x}) \rbrack + \mathds{E}_{\bm{\hat{x}} \sim \mathds{P}_g} \lbrack \log (1 - D(\bm{\hat{x}})) \rbrack \label{eq:standard_gan} 	
\end{equation} We refer to this value function as $V(D,G)$. The loss function for the discriminator and generator are respectively defined as: 
	\begin{align}
	\begin{split}\label{eq:ganD_loss_standard}
	\mathcal{L}^{GAN}_{D} &= - \mathds{E}_{\bm{x} \sim \mathds{P}_r} \lbrack \log D(\bm{x}) \rbrack - \mathds{E}_{\bm{\hat{x}} \sim \mathds{P}_g} \lbrack \log (1 - D(\bm{\hat{x}})) \rbrack
	\end{split}\\
	\begin{split}\label{eq:ganG_loss_standard}
		\mathcal{L}^{GAN}_{G} &=  \mathds{E}_{\bm{\hat{x}} \sim \mathds{P}_g} \lbrack \log (1 - D(\bm{\hat{x}})) \rbrack
	\end{split}
	\end{align}
In practice we train $G$ to maximize $\log(D(\bm{\hat{x}}))$, because at the start of training the generated samples $\bm{\hat{x}}$ are of poor quality, making it easy for the discriminator to distinguish $\mathds{P}_g$ from $\mathds{P}_r$. This could saturate $\log (1 - D(\bm{\hat{x}}))$. Both the generator and the discriminator must be differentiable functions. In section \ref{sec:experiments} we will go into more depth regarding the used networks that represent $G$ and $D$.

\subsection{Conditional GAN}\label{sec:conditional}
We can extend the standard GAN such that we can generate samples conditioned on some variable \citep{conditional_gan}. This can be very helpful when we want to generate samples of a specific mode.  This setup is referred to as conditional GAN, and works as follows. We include an additional variable $\bm{y}$ in the model, such that we generate and discriminate samples conditioned on $\bm{y}$. The goal for the generator is then to generate samples $\bm{\hat{x}}$ that are both realistic and match the context $\bm{y}$. It is possible to use various types of data for $\bm{y}$, such as discrete labels \citep{conditional_gan}, text \citep{text_to_image} or even images \citep{image_to_image}. One of the benefits of a conditional GAN is that it is able to learn improved representations for multi-modal data. The objective function for a conditional GAN is: \begin{equation}\begin{aligned} \min_{G} \max_{D} \mathds{E}_{\bm{x},\bm{y} \sim \mathds{P}_r} \lbrack \log D(\bm{x} , \bm{y}) \rbrack  +  \mathds{E}_{\bm{z}\sim p(\bm{z}),\bm{y}\sim \mathds{P}_y} \lbrack \log (1 - D(G(\bm{z},\bm{y}) , \bm{y})) \rbrack \label{eq:conditional_gan}\end{aligned} \end{equation} where $\mathds{P}_y$ is the conditional density distribution. In the conditional GAN, we concatenate the conditional variable $\bm{y}$ to the input of every layer, for both the generator and discriminator. A conditional GAN is a supervised network that needs for each data point $\bm{x}$ a label $\bm{y}$. This network has many similarities with InfoGAN \citep{infogan}, an unsupervised conditional GAN, which we will introduce in the next section.
%TODO: improve notation, and explain how it is implemented

\subsection{InfoGAN}\label{sec:infoGAN}
In the standard GAN setup (as described in section \ref{sec:GAN}), there is no reason to expect that individual dimensions of the latent vector correspond to specific semantic features in the generated data. Preliminary research has shown that varying a single dimension of the latent vector has very little influence on the generated images.
In section \ref{sec:conditional} we discussed a supervised method to train a GAN, where we ideally can change visual features of the generated data, by changing the conditional variable $\bm{y}$. However, in many cases we have sparse labels, or no labels at all for the given dataset. \cite{infogan} proposed to use an additional latent code $\bm{c}$, that is used together with $\bm{z}$ as an input for the generator. In contrast to the conditional GAN, the discriminator does not know about $\bm{c}$ (or $\bm{y}$ in the case of the conditional GAN). Because the discriminator has no information about $\bm{c}$, the easiest solution for the generator would be to ignore $\bm{c}$ completely. In order to prevent this, \cite{infogan} proposed to add an additional term to the loss functions that promotes high mutual information between $\bm{c}$ and $G(\bm{z},\bm{c})$. Mutual information between $X$ and $Y$ is denoted as $I(X; Y)$, and measures the reduction of uncertainty in $Y$ when $X$ is observed, and vice versa. The minimax game we try to solve in InfoGAN is then the same as Equation \ref{eq:standard_gan}, with $-I(\bm{c};G(\bm{z},\bm{c}))$ added.	The posterior $P(\bm{c}\vert \bm{\hat{x}})$, needed to determine $I(\bm{c};G(\bm{z},\bm{c}))$, is intractable to calculate. Therefore we make use of an additional distribution $Q(\bm{c}\vert\bm{\hat{x}})$ that approximates $P(\bm{c}\vert \bm{\hat{x}})$. \cite{infogan} show that you can maximize the following lower bound in order to maximize $I(\bm{c};G(\bm{z},\bm{c}))$: \begin{equation} \label{eq:lower_bound}
	I(\bm{c};G(\bm{z},\bm{c})) \geq \mathds{E}_{\bm{c} \sim \mathds{P}(\bm{c}),\bm{\hat{x}} \sim G(\bm{z},\bm{c})} \lbrack \log Q(\bm{c}\vert\bm{\hat{x}}) \rbrack  \end{equation} We refer to this lower bound as $L_I(G,Q)$, where $Q$ is a neural network that shares all layers with $D$ (except the last one). The last layer of $Q$ gives as an output the conditional distribution $Q(\bm{c}\vert\bm{\hat{x}})$. The minimax game for InfoGAN is as follows: \begin{equation}\label{eq:infogan}
		\min_{G,Q} \max_{D} V_{\text{InfoGAN}}(D,G,Q) = V(D,G) - \lambda_{\text{I}} L_I(G,Q)
	\end{equation} where $\lambda_{\text{I}}$ is a hyperparameter.
	The latent variable $\bm{c}$ can consist of multiple separate distributions. In this research we focus on categorical and continuous latent codes. The first is represented by a softmax nonlinearity in $Q$, the latter is represented by a factored Gaussian.
	In section \ref{sec:experiments} we will provide the used latent codes per dataset.

\subsection{Wasserstein GAN}
GANs are known to be very difficult to train for various reasons \citep{training_gans}. With the original value function as defined in \ref{eq:standard_gan}, a strong discriminator can cause vanishing gradients. With the improved value function, where the generator maximizes $\log(D(\hat{\bm{x}}))$, the training updates can be very unstable. In both situations the generator may not converge to a stable network that produces realistic samples. %TODO: explain mode colapse
In the work of \cite{wasserstein} an alternative value function named Wasserstein GAN (WGAN) is proposed. The authors show that the value function of WGAN has better convergence properties compared to the standard value function. WGAN makes use of the Wasserstein distance $W(q,p)$, which intuitively can be interpreted as the minimum cost of moving mass such that distribution $q$ is transformed to distribution $p$ (where cost is the transformed mass multiplied with the distance it has been moved). The WGAN value function is as follows: \begin{equation}\label{eq:wgan}
	\min_G \max_{D\in\mathcal{D}} \mathds{E}_{\bm{x} \sim \mathds{P}_r} \lbrack D(\bm{x}) \rbrack - \mathds{E}_{\bm{\hat{x}}\sim \mathds{P}_g} \lbrack D(\bm{\hat{x}}) \rbrack \end{equation}
where $\mathcal{D}$ is the set of functions that are 1-Lipschitz\footnote{A function $f$ is 1-Lipschitz if $\vert f(a) - f(b) \vert \leq \vert a - b \vert$ for all $a$ and $b$ in the domain of $f$.}. Because the discriminator in the WGAN does not discriminate anything (e.g. real images from fake images), it is referred to as \textit{critic}. With an optimal critic, the generator minimizes $W(\mathds{P}_r,\mathds{P}_g)$ when we minimize the value function from Equation \ref{eq:wgan}. In order to enforce the Lipschitz constraint, \cite{wasserstein} make use of clipping the weights of the critic between $\lbrack -c,c \rbrack$ for some positive constant $c$. However, \cite{improved_wasserstein} show that weight clipping can result in undesired behaviour such as exploding or vanishing gradients.
\cite{improved_wasserstein} propose to use a gradient penalty to enforce the Lipschitz constraint, this method is referred to as WGAN with gradient penalty (WGAN-GP). The gradient penalty is defined as \begin{equation} \label{eq:gradient_penalty}\mathds{E}_{\bm{\tilde{x}}\sim \mathds{P}_{\bm{\tilde{x}}}} \lbrack ( \| \nabla_{\bm{\tilde{x}}} D(\bm{\tilde{x}}) \|_2 - 1)^2 \rbrack \end{equation} where $\mathds{P}_{\bm{\tilde{x}}}$ is the distribution sampled uniformly along straight lines between points in the distributions $\mathds{P}_r$ and $\mathds{P}_g$ (see line 8 in Algorithm \ref{alg:wgan_gp}). Because the gradient penalty is determined for each sample independently (see Algorithm \ref{alg:wgan_gp}), we omit batch normalization in the critic. Batch normalization would namely create dependence between the samples within a batch. We should train the critic till optimality, but in practice we train the critic for a specific amount of iterations indicated by $n_{\text{critic}}$. The coefficient $\lambda_{\text{GP}}$, shown on line 9 in Algorithm \ref{alg:wgan_gp}, is used to make sure the effect of the gradient penalty is significant. 

\begin{algorithm}[!tbp] 
\caption{WGAN-GP}
\label{alg:wgan_gp}
\begin{algorithmic}[1]
\REQUIRE gradient penalty coefficient $\lambda_{\text{GP}}$, number of critic updates per generator update $n_{\text{critic}}$, batch size $m$
\REQUIRE initial critic parameters $w_0$, initial generator parameters $\theta_0$
\WHILE{not converged}
\FOR{$t=1,...,n_{\text{critic}}$}
\FOR{$i=1,...,m$}
\STATE $\bm{x} \sim \mathds{P}_r$
\STATE $\bm{z} \sim p(\bm{z})$
\STATE $\epsilon \sim U\lbrack 0,1 \rbrack$
\STATE $\bm{\hat{x}} \leftarrow G_\theta(\bm{z})$
\STATE $\bm{\tilde{x}} \leftarrow \epsilon \bm{x} + (1-\epsilon)\bm{\hat{x}}$
\STATE $GP \leftarrow \lambda_{\text{GP}} \left( \| \nabla_{\bm{\hat{x}}} D_w(\bm{\tilde{x}}) \|_2 -1 \right)^2$ 
\STATE $\mathcal{L}^{(i)} \leftarrow D_w(\bm{\hat{x}}) - D_w(\bm{x}) + GP$
\ENDFOR
\STATE $w \leftarrow \text{Adam}(\nabla_w \frac{1}{m}\sum_{i=1}^m \mathcal{L}^{(i)}, w)$
\ENDFOR
\STATE $\lbrace\bm{z}\rbrace_{i=1}^m \sim p(\bm{z})$
\STATE $\theta \leftarrow \text{Adam}(\nabla_\theta \frac{1}{m} \sum_{i=1}^m - D_w( G_\theta (\bm{z})), \theta)$
\ENDWHILE
\end{algorithmic}
\end{algorithm}

\subsection{Encoder}\label{sec:encoder}
\begin{figure}[H]
	\centering
	\includegraphics[width=.8\linewidth]{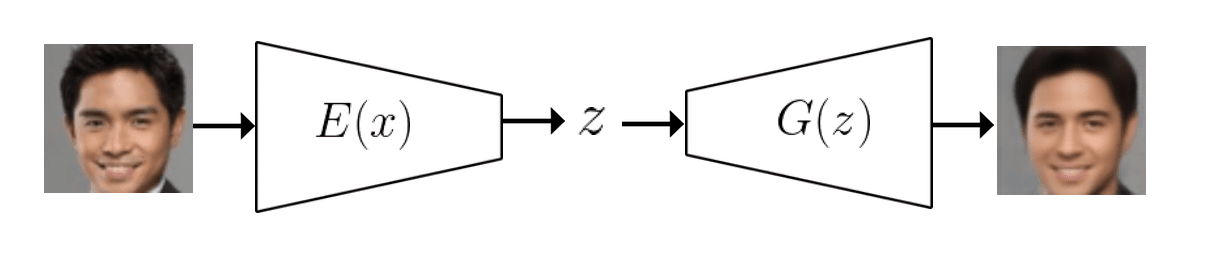}
	\caption{Graphical representation of the encoder in a GAN.}
	\label{img:encoder}
\end{figure}

In the standard GAN setup we can create random samples that seem to originate from the real data distribution $\mathds{P}_r$. However, we can not create samples that look similar to some specific data points in $\mathds{P}_r$. Formally, we would like to have a method such that we can find a latent vector $\bm{\hat{z}}$ for a sample $\bm{x}$, such that $G(\bm{\hat{z}}) \approx \bm{x}$ (as illustrated in Figure \ref{img:encoder}). This method has many similarities with a variational autoencoder \citep{kingma2013auto}, where the decoder is replaced by the generator. In order to find such a latent vector, we will use an encoder $E$ that has as objective $G(E(\bm{x})) \approx \bm{x}$ for all $\bm{x} \sim \mathds{P}_r$. In order to train the encoder we can make use of several loss functions \citep{loss_functions}. In this research we focused on minimizing the following loss function: \begin{equation*} \label{eq:encoder_loss} \mathcal{L}_{E} = \mathds{E}_{\bm{x}\sim \mathds{P}_r} \lbrack \frac{1}{N} \sum_{p \in P} \vert G(E(\bm{x}))(p) - \bm{x}(p) \vert  \rbrack \end{equation*} where $N$ is the number of pixels, $p$ is the index of a pixel, and $P$ is the set of indices of all pixels. Note that when using either the condition GAN or infoGAN, we need to encode an additional variable ($\bm{y}$ or $\bm{c}$ respectively). Similar to the generator and discriminator, the encoder is parameterized by a neural network. %TODO: add sources 

\subsection{Capsule Network}\label{sec:capsule}
In the original paper that introduced GANs \citep{generative} the authors use a multilayer perceptron for both the generator and discriminator. Although this network performs reasonably well on a simple dataset (e.g. MNIST), on more complex datasets the results are not good. Convolutional neural networks (CNNs) are widely used for supervised learning in computer vision problems. In more recent work \citep{dcgan} the GAN setup was succesfully combined with a CNN (DCGAN) to create state of the art results. \cite{dcgan} proposed several improvements such as using strided convolutions, batch normalization \citep{batch_normalization}, and the ReLU activation function. %TODO: add citiations
Although CNNs are good at detecting local features, the technique is less effective at detecting (spatial) relationships between these different features. Part of this problem is caused by the invariant detection of features, CNNs process the likelihood of features without subsequently processing the properties of these features (e.g. angle or size). \cite{capsules} propose a network that uses capsules to present an object or features of an object. The activity of these capsules is represented by a vector, where the length of the vector represents the likelihood of an object or part of an object. The orientation of the vector represents the specific properties of this object. When using multiple layers of capsules, the predictions of higher-level capsules are determined by the lower-level capsules and a transformation matrix. Because lower-level capsules also capture the instantiation of parts of an object, high-level capsules can make better predictions about the global features of an object.

We will now describe the capsule network formally. Let $\bm{s}_j$ denote the input vector of capsule $j$, then the output vector is defined as \begin{equation}\label{eq:squash} \bm{v}_j = \mathtt{squash}(\bm{s}_j) =  \frac{\| \bm{s}_j \|^2}{1 + \| \bm{s}_j \|^2} \frac{\bm{s}_j}{\| \bm{s}_j \|} \end{equation} Note that the orientation of the vector is preserved, while short vectors are reduced to almost zero length, and long vectors are reduced to approximately unit length. For a capsule $j$ the total input is a weighted sum of all the vectors $\bm{\hat{u}}_{j \vert i}$, where $\bm{\hat{u}}_{j \vert i}$ is a prediction vector of capsule $i$ (in the layer below) connected to capsule $j$, calculated as \begin{equation} \label{eq:caps_layer}
	\bm{\hat{u}}_{j \vert i} = \bm{W}_{ij} \bm{v}_i \end{equation} where $\bm{W}_{ij}$ is a weight matrix and $\bm{v}_i$ the output of a capsule in the layer below. The total input for capsule $j$ is then calculated as follows \begin{equation} \bm{s}_j = \sum_i c_{ij} \bm{\hat{u}}_{j \vert i} \end{equation} where $c_{ij}$ are coupling coefficients. The coupling coefficients are determined by a technique called "routing-by-agreement" \citep{capsules}, such that the coefficients are increased when the prediction vector is similar to the output of a parent capsule, and decreased when they are dissimilar. Similarity between two vectors is measured as the scalar product of these two vectors. The coupling coefficients of capsule $j$ and all the parent capsules sum to 1, enforced by \begin{equation}\label{eq:softmax} c_{ij} = \frac{\exp(b_{ij})}{\sum_k \exp(b_{ik})} \end{equation} where $b_{ij}$ are the log prior probabilities that capsules $i$ and $j$ should be connected. The coupling coefficients are refined multiple times to make connections between agreeing capsules stronger, as described in Algorithm \ref{alg:routing}.

\begin{algorithm}[!tbp] 
\caption{Dynamic Routing}
\label{alg:routing}
\begin{algorithmic}[1]
\REQUIRE prediction vectors $\bm{\hat{u}}_{j \vert i}$, routing iterations $r$, capsule layer $l$
\REQUIRE $b_{ij}$ initial routing logits for all capsules $i$ in layer $l$ and capsules $j$ in layer $(l+1)$
\FOR{$r$ iterations}
\FOR{all capsules $i$ in layer $l$}
\STATE $\bm{c}_i \leftarrow \mathtt{softmax}(\bm{b}_i)$ \COMMENT{$\triangleright$ Eq. \ref{eq:softmax}}
\ENDFOR
\FOR{all capsules $i$ in layer $l+1$}
\STATE $\bm{s}_j \leftarrow \sum_i c_{ij} \bm{\hat{u}}_{j \vert i}$
\ENDFOR
\FOR{all capsules $j$ in layer $l+1$}
\STATE $\bm{v}_j \leftarrow \mathtt{squash}(\bm{s}_j)$ \COMMENT{$\triangleright$ Eq. \ref{eq:squash}}
\ENDFOR
\FOR{all capsules $i$ in layer $l$}
\FOR{all capsules $j$ in layer $l+1$}
\STATE $b_{ij} \leftarrow b_{ij} + \bm{\hat{u}}_{j \vert i} \cdot \bm{v}_j$
\ENDFOR
\ENDFOR
\ENDFOR
\end{algorithmic}
\end{algorithm}

\section{Experiments}\label{sec:experiments}
In our experiments we make use of two datasets, MNIST \citep{mnist} and CelebA \citep{celebA}. The discussed techniques are very general, making them applicable to a broad range of image datasets. We use two datasets to compare the techniques on relatively simple, and more complex images.

First of all, we want to generate realistic looking images. We will compare the generated images when different discriminators are used during training. By means of the conditional GAN we will experiment with generating images with certain specified attributes. Using the encoder network, we aim to perfectly reconstruct the images from the dataset. Finally, by combining a conditional GAN with the encoder network, we will experiment with changing visual attributes of reconstructed images.

\subsection{MNIST}\label{sec:mnist}
The first dataset we used is MNIST \citep{mnist}, consisting of 60,000 1-channel images of handwritten digits with a size of $28\times28$ pixels. For supervised learning (conditional GAN), we can make use of the image labels indicating the digit. %TODO: rephrase last sentence

\subsection{CelebA face images}\label{sec:celebA}
The second dataset we used to compare the techniques is CelebA \citep{celebA}. This dataset consists of 202,599 images of faces, every image is labelled with 40 binary attributes. Examples of these attributes are smiling, wearing hat, mustache, and blond hair. The images are rescaled and cropped to images of $64\times64$ pixels. The original size of the images is $178\times218$ pixels.

\subsection{Architectures}\label{sec:architectures}
We will now discuss the architectures used in this research. The generator and discriminator are similar to the convolutional neural networks proposed by \cite{dcgan}. As described in Table \ref{table:networks} we use the same encoder for both datasets. For the generator we found that the results are best if we use different networks for the different datasets. In the generator we use transposed convolutional layers with stride 2, in order to upscale the images.
In all experiments we used the Adam optimizer \citep{adam} to minimize the loss functions, with a learning rate of $10^{-4}$ and batch size of 64. For the Adam optimizer we used 0.5 for $\beta_1$ and $0.9$ for $\beta_2$ when training WGAN-GP, otherwise we used 0.5 and 0.99 respectively. For WGAN-GP we used a gradient penalty  coefficient of 10 and 5 critic updates per generator update, similar to \cite{wasserstein}. These parameters are indicated in Algorithm \ref{alg:wgan_gp} by $\lambda_{\text{GP}}$ and $n_{\text{critic}}$, respectively. We trained all models for 60 epochs. In all experiments with the MNIST dataset we used a $\bm{z}$ vector of dimension 64, whereas with the CelebA dataset we used vectors of dimension 128. In all experiments the latent vector follows the uniform distribution $\mathcal{U}(-1,1)$. When training infoGAN the additional $Q$ network shares all layers (except the last one) with the discriminator. The final output for the conditional distribution $Q(\bm{c}\vert\bm{\hat{x}})$ is determined by using two fully-connected layers, of which the first one has 128 hidden nodes. The second layer has an output dimension that matches the predetermined dimension of the latent code $\bm{z}$. When we train a conditional GAN or infoGAN, the output dimension of network $E$ is equivalent to the total dimensions of the latent vector and vectors $\bm{y}$ or $\bm{c}$. Note that in this context the latent vector $\bm{z}$ is not really a noise vector.

In the experiments with the capsule network as discriminator we used the following parameters. The network is similar to the one proposed by \cite{capsules}, with the only differences that the first layer has 128 filters instead of 256, and that the final layer consists of only a single capsule instead of 10 capsules. Our network has a single output capsule because the task is binary classification (differentiate real from generated images), whereas \cite{capsules} used 10 output capsules because they tried to classify digits in the MNIST dataset. The first layer is a standard convolutional layer with 128 convolutional filters with size $9\times9$, a stride of 1, and the ReLU activation function. In the second layer we use a convolutional capsule layer with 32 channels, where each channel is an 8 dimensional capsule. This 8 dimensional capsule has filters of size $9\times9$ and a stride of 2. Using dynamic routing (as described in Algorithm \ref{alg:routing}), we map the output of the second layer to the third and also final layer. For the dynamic routing algorithm we use 3 routing iterations. The final layer consists of a single 16 dimensional capsule. \cite{capsules} use an additional reconstruction loss to promote regularization, but in our experiments we omitted this term. 

\begin{table*}[h]
\centering
\caption{Architectures for the encoder, generator and discriminator. FC stands for a fully-connected layer, BN denotes batch normalization, and t-conv is short for transposed convolution. $\bm{z}$-dim is 64 when using MNIST, for CelebA it is 128. All the convolutional filters are of size $5\times5$, with stride 2. Note that we use different generator networks for the two datasets. lRelU is short for leaky ReLU, for this activation function we used a slope coefficient of 0.2. }
\label{table:networks}
\begin{tabular}{@{}llll@{}}
\toprule
\multicolumn{1}{c}{Encoder} & \multicolumn{1}{c}{Generator MNIST}    & \multicolumn{1}{c}{Generator CelebA}   & \multicolumn{1}{c}{Discriminator} \\ \midrule
64 conv. lReLU          & 7$\times$7$\times$128 FC, BN, ReLU   & 4$\times$4$\times$512 FC, BN, ReLU   & 64 conv. lReLU                \\
128 conv. BN, lReLU  & 128 t-conv. BN, ReLU & 256 t-conv. BN, ReLU & 128 conv. BN, lReLU        \\
256 conv. BN, lReLU  & 1 t-conv. tanh                     & 128 t-conv. BN, ReLU & 256 conv. BN, lReLU        \\
512 conv.  BN, lReLU &                                        & 64 t-conv. BN, ReLU  & 512 conv. BN, lReLU        \\
$\bm{z}$-dim FC, tanh   &                                        & 3 t-conv. tanh                     & 1 FC, sigmoid        \\ \bottomrule
\end{tabular}
\end{table*}

\subsection{Results}
The shown results are from the first experiments after hyperparameter tuning. We ran multiple experiments to make sure the result are similar in different runs, ensuring that the results are representative.

\begin{figure}[H]
\captionsetup{width=0.4\textwidth}
\centering
\begin{minipage}{.5\textwidth}
 \centering
	\includegraphics[width=.8\linewidth]{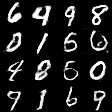}
	\captionof{figure}{Generated samples for the MNIST dataset. The used networks are described in Table \ref{table:networks}, trained with WGAN-GP.}
	\label{img:mnist_random}
\end{minipage}%
\begin{minipage}{.5\textwidth}
  \centering
	\includegraphics[width=.8\linewidth]{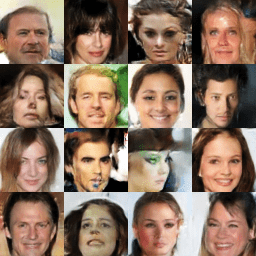}
	\captionof{figure}{Generated samples for the CelebA dataset. The used networks are described in Table \ref{table:networks}, trained with WGAN-GP.}
	\label{img:celebA_random}
\end{minipage}
\end{figure}

We found that using the WGAN-GP training, the results are of higher quality compared to the standard GAN training objective. In Figures \ref{img:mnist_random} and \ref{img:celebA_random} you can see generated samples for the two datasets, using the network as described in Table \ref{table:networks}, trained with WGAN-GP. 

\begin{figure}[H]
\captionsetup{width=0.4\textwidth}
\centering
\begin{minipage}{.5\textwidth}
 \centering
	\includegraphics[width=.8\linewidth]{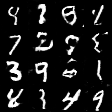}
	\caption{Generated samples for the MNIST dataset, trained using the capsule network as discriminator. The generator is described in Table \ref{table:networks}.}
	\label{img:mnist_random_capsule}
\end{minipage}%
\begin{minipage}{.5\textwidth}
  \centering
	\includegraphics[width=.8\linewidth]{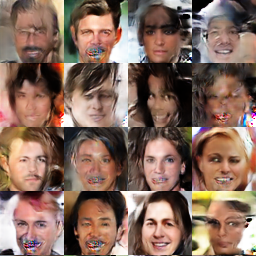}
	\caption{Generated samples for the CelebA dataset, trained using the capsule network as discriminator. The generator is described in Table \ref{table:networks}.}
	\label{img:celebA_random_capsule}
\end{minipage}
\end{figure}

\begin{figure}[H]
\captionsetup{width=0.44\textwidth}
\centering
\begin{minipage}{.5\textwidth}
 \centering
	\includegraphics[width=.8\linewidth]{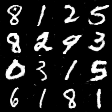}
	\caption{Generated samples for the MNIST dataset. The used networks are described in Table \ref{table:networks}, trained using the standard GAN objective.}
	\label{img:mnist_random_dcgan}
\end{minipage}%
\begin{minipage}{.5\textwidth}
  \centering
	\includegraphics[width=.8\linewidth]{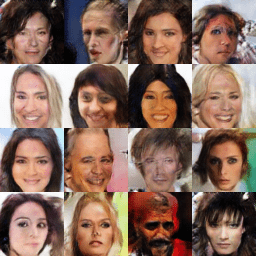}
	\caption{Generated samples for the CelebA dataset. The used networks are described in Table \ref{table:networks}, trained using the standard GAN objective.}
	\label{img:celebA_random_dcgan}
\end{minipage}
\end{figure}

In Figures \ref{img:mnist_random_capsule} and \ref{img:celebA_random_capsule} the results of using the capsule network as a discriminator are shown. For the training of this network we used the standard GAN objective (Equation \ref{eq:standard_gan}). We did not manage to combine the capsule network with the WGAN-GP objective, because determining the gradient penalty (Equation \ref{eq:gradient_penalty}) is nontrivial when using dynamic routing (Algorithm \ref{alg:routing}). Figures \ref{img:mnist_random_dcgan} and \ref{img:celebA_random_dcgan} show the results of using standard convolutional network (as described in Table \ref{table:networks}) with the standard GAN objective. The generated samples using the capsule network are of a lower quality compared to the samples generated when using a standard convolutional network as discriminator. 

\begin{figure}[H]
	\centering
	\includegraphics[width=.6\linewidth]{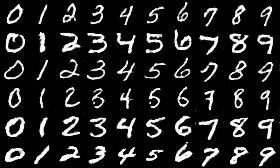}
	\caption{Random samples for MNIST, conditioned on digit label. Every column corresponds to a unique label, the samples in a row have the same latent noise vector. The used networks are described in Table \ref{table:networks}, trained with the conditional WGAN-GP.}
	\label{img:mnist_conditional}
\end{figure}

\begin{figure}[H]
	\centering
	\includegraphics[width=.6\linewidth]{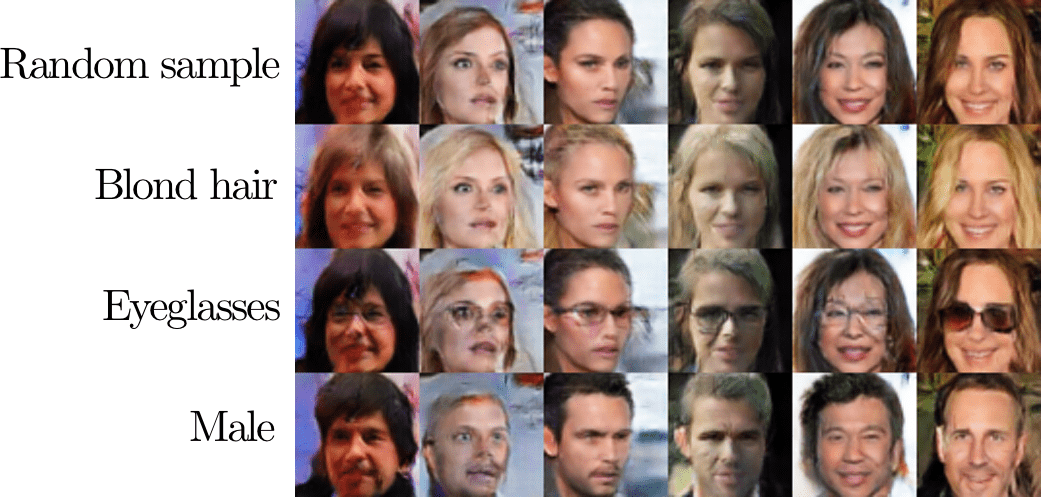}
	\caption{Random samples for CelebA, conditioned on three different attributes. Every column corresponds to a unique latent noise vector. The used networks are described in Table \ref{table:networks}, trained with the conditional WGAN-GP.}
	\label{img:celebA_conditional}
\end{figure}

For the conditional GAN we performed the following experiments. When training the conditional GAN we could use either the standard GAN objective or the WGAN-GP objective. We found that, similar to the unconditional GAN, the results are better when using WGAN-GP. In Figure \ref{img:mnist_conditional} the results for the MNIST dataset are shown. For the conditional variable we used the digit classes. The results show that this supervised network makes it possible to generate images of distinct classes, where the style of these images is determined by the latent noise vector. The results of the conditional GAN for the CelebA dataset are shown in Figure \ref{img:celebA_conditional}, where we made use of the binary attributes blond hair, eyeglasses, and male. The results show that by varying the conditional variables, we can visually change the sample generated from the same noise vector. Figures \ref{img:celebA_reconstruct_conditional} and \ref{img:celebA_conditional_wgan} show more results for the conditional GAN using the CelebA dataset.

\begin{figure}[H]
	\centering
	\includegraphics[width=.5\linewidth]{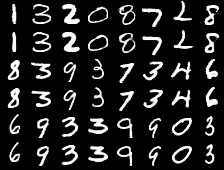}
	\caption{Image encoding and reconstruction for the MNIST dataset. Rows 1, 3, and 5 are the original images taken from MNIST, rows 2, 4, and 6 are the corresponding reconstructed images based on the encoding. The used networks are described in Table \ref{table:networks}, trained with WGAN-GP.}
	\label{img:mnist_encoding}
\end{figure}

\begin{figure}[H]
	\centering
	\includegraphics[width=.6\linewidth]{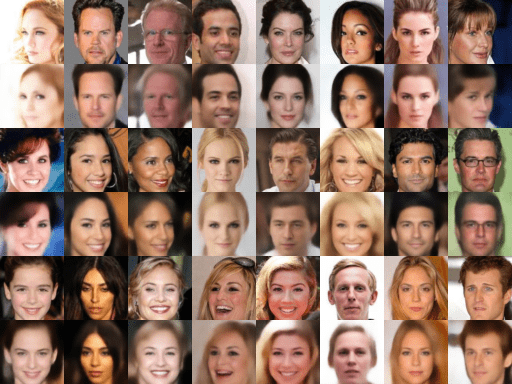}
	\caption{Image encoding and reconstruction for the CelebA dataset. Rows 1, 3, and 5 are the original images taken from CelebA, rows 2, 4, and 6 are the corresponding reconstructed images based on the encoding. The used networks are described in Table \ref{table:networks}, trained with WGAN-GP.}
	\label{img:celebA_encoding}
\end{figure}

In order to generate images that look similar to a specific image we made use of the encoder network. Figures \ref{img:mnist_encoding} and \ref{img:celebA_encoding} show the results of the encoding and reconstruction for MNIST and CelebA, respectively. For MNIST we find that the encoding results in very similar pictures. In many images it is difficult to distinguish the real images from their encoded and reconstructed counterparts. The encoding of the CelebA images shows that high frequency details are lost. However, the encoded images still show a great similarity with the original images. 

\begin{figure}[H]
	\centering
	\includegraphics[width=.6\linewidth]{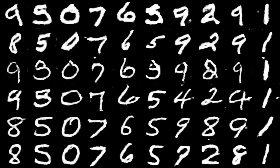}
	\caption{Manipulation of categorical latent variable $\bm{c}$ for MNIST samples. Every column corresponds to a unique category, every row has a fixed latent vector. The used networks are described in Table \ref{table:networks}, trained with InfoGAN.}
	\label{img:mnist_infogan_cat}
\end{figure}

\begin{figure}[H]
	\centering
	\includegraphics[width=.6\linewidth]{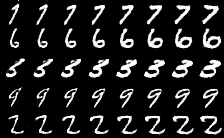}
	\caption{Manipulation of continuous latent variable $\bm{c}$ for MNIST samples. Every row corresponds to a unique latent variable, where we vary $\bm{c}$ between -1 and 1. The used networks are described in Table \ref{table:networks}, trained with InfoGAN.}
	\label{img:mnist_infogan_cont}
\end{figure}

\begin{figure}[H]
	\centering
	\includegraphics[width=.6\linewidth]{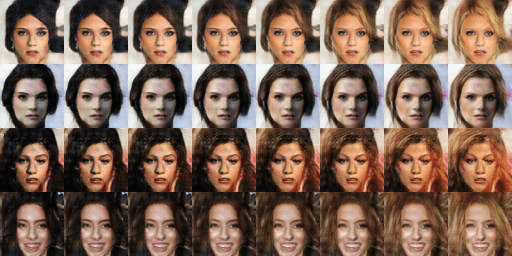}
	\caption{Manipulation of continuous latent variable $\bm{c}$ for CelebA samples. The variable is varied between -1 and 1, every row has a unique latent vector. The used networks are described in Table \ref{table:networks}, trained with InfoGAN.}
	\label{img:celebA_infogan}
\end{figure}

We will now compare some results for the infoGAN setup. For the MNIST dataset we experimented with using a categorical variable with 10 classes, in the hope that every class corresponds to a unique digit. Additionally, we used a continuous variable following a uniform distribution. The results for this experiment are illustrated in Figures \ref{img:mnist_infogan_cat} and \ref{img:mnist_infogan_cont}, respectively. By varying the categorical variable, we observe a change in digit class. However, we see that for digits that look similar (e.g. 5 and 9), a single categorical class can contain both digit classes. Varying the continuous latent variable influences the width of the generated digits. In Figure \ref{img:celebA_infogan} the results are shown for varying a continuous latent variable on the CelebA dataset. We observe that this variable has an influence on hair colour, changing the colour from dark to bright.


\section{Conclusion and Future Research}\label{sec:conclusion}
In this paper we compared several GAN techniques for image creation and modification. We found that the standard GAN can be unstable, using WGAN-GP results in more stable training and the generated images are of a higher quality. With the standard GAN, we often observed a mode collapse. The mode collapse usually occurred after many epochs, making training very ineffective. It can take a lot of hyperparameter tuning to find the correct balance between the generator and discriminator, making sure one doesn't outperform the other. When using WGAN-GP, this balance is less of a concern because the critic is trained till optimality.   

We compared two conditional GANs, namely the standard conditional GAN that is supervised and InfoGAN, an unsupervised network. The supervised conditional GAN makes it possible to vary specific visual attributes within an image, given that the dataset contains labels for these attributes. With infoGAN, it is also possible to change visual attributes of the generated images. However, in contrast to the supervised conditional GAN, it is not possible to specify which visual attributes we would like to change. Depending on the dataset and the chosen latent distribution, the network learns a disentangled representation. We also used an encoder network that makes it possible to create a reconstruction of an image, independent of the used GAN variant. The reconstructions for MNIST are seemingly perfect, whereas for the CelebA dataset the reconstructions are good but blurry. Using this technique we can apply dimensionality reduction, similar to a variational autoencoder \citep{kingma2013auto}. We experimented with using the novel capsule network as discriminator. This however did not lead to satisfying results. We expect that this is due to the relatively small and shallow network.\\ 

Because of the rapid development of GAN techniques, many opportunities for future research are remaining. In follow-up research we would like to experiment with reconstructing images using gradient based approaches to generate latent vectors, similar to \cite{lipton2017precise} their approach. Furthermore, more experiments with different loss functions for the encoder network are needed. We could even extend this, using the discriminator in a GAN as a measure for the reconstruction objective, as introduced by \cite{autoencoding_gan}. More experiments with methods that improve the stability of GANs are needed, such as spectral normalization \citep{spectral} and using two discriminators \citep{dual_discriminator}. Finally, we want to test the discussed methods on different and more complex datasets.
\section{Acknowledgements}\label{sec:acknowledgements}
We would like to thank the Center for Information Technology of the University of Groningen for their support and for providing access to the Peregrine high performance computing cluster.

\bibliographystyle{plainnat}
\bibliography{literature}

\clearpage

\begin{appendices}
	\section{Appendix}\label{sec:appendix_A}
\subsection{Bilinear Interpolation on Latent Space}
\begin{figure}[h]
	\centering
	\includegraphics[width=.45\linewidth]{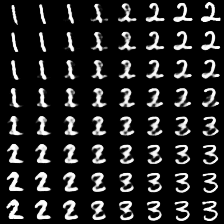}
	\caption{Bilinear interpolation on the latent space for encoded MNIST samples. The used networks are described in Table \ref{table:networks}, trained with WGAN-GP.}
	\label{img:mnist_interpolate_wgan}
\end{figure}

\begin{figure*}[h]
	\centering
	\includegraphics[width=1\linewidth]{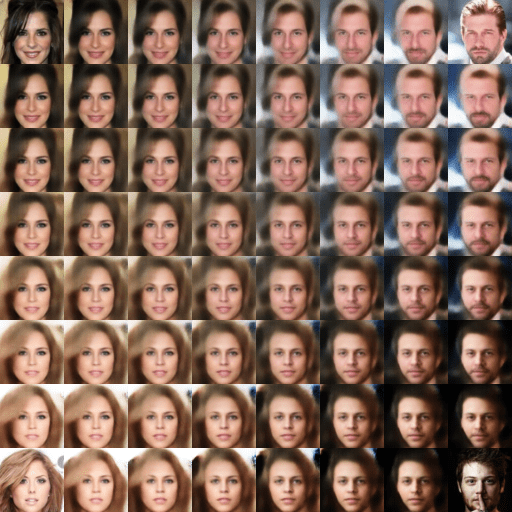}
	\caption{Bilinear interpolation on latent space for encoded CelebA samples. The used networks are described in Table \ref{table:networks}, trained with WGAN-GP.}
	\label{img:celebA_interpolate_wgan}
\end{figure*}

\begin{figure*}[h]
	\centering
	\includegraphics[width=1\linewidth]{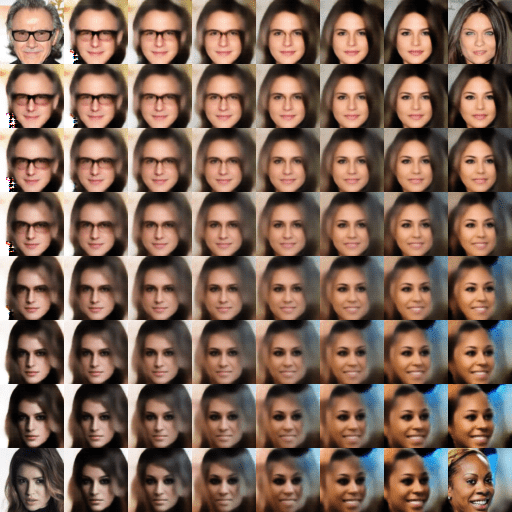}
	\caption{Bilinear interpolation on latent space for encoded CelebA samples. The used networks are described in Table \ref{table:networks}, trained with WGAN-GP.}
	\label{img:celebA_interpolate_wgan1}
\end{figure*}

\begin{figure*}[h]
	\centering
	\includegraphics[width=1\linewidth]{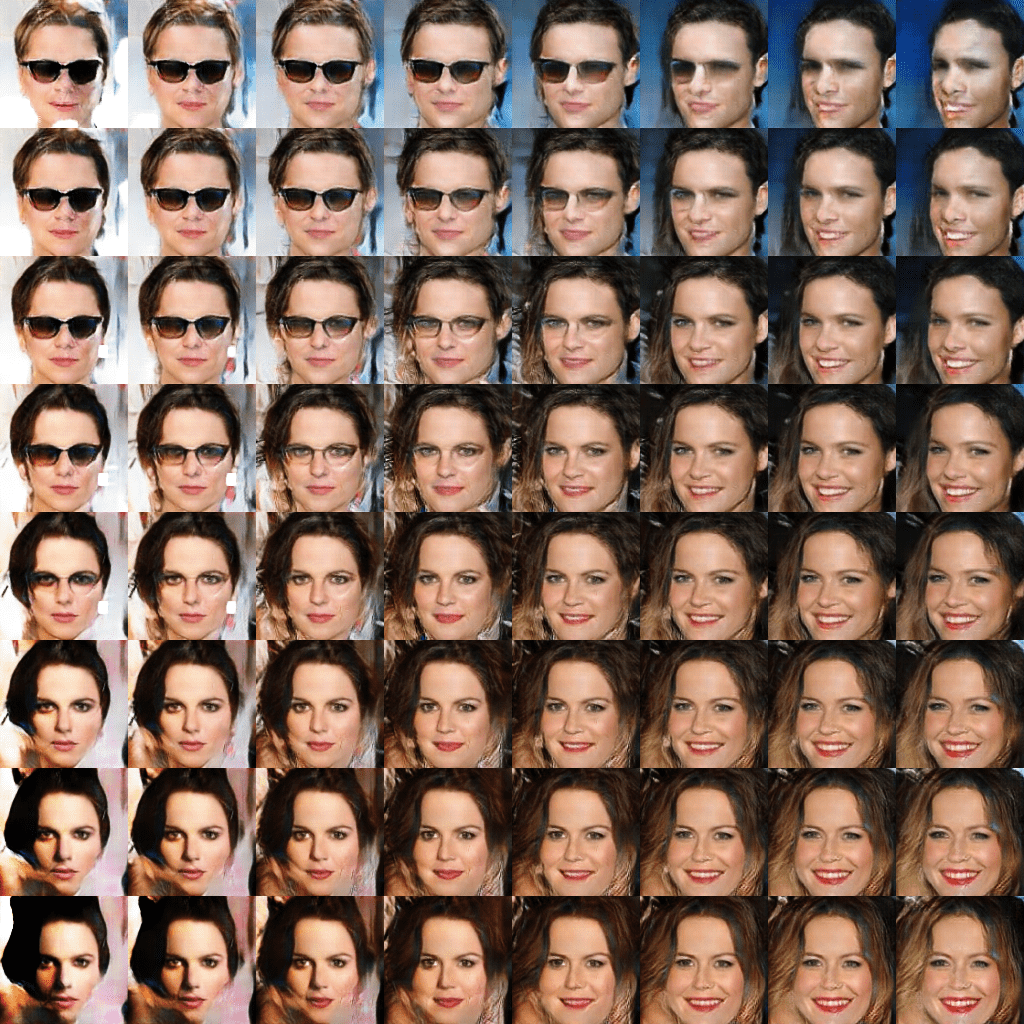}
	\caption{Bilinear interpolation on latent space for random noise vectors. Dataset used is CelebA with images of size 128$\times$128. The used networks are described in Table \ref{table:networks}, trained with WGAN-GP.}
	\label{img:celebA_interpolate128}
\end{figure*}

In order to show the process of encoding images as well as applying interpolation between images we use the following process. Every corner of a figure contains a real image from the given dataset. We use the encoder network to generate four latent vectors for the given images, and subsequently apply bilinear interpolation between these four vectors. These vectors are then used as an input for the generator, producing the final images. An example for bilinear interpolation on the MNIST dataset is given in Figure \ref{img:mnist_interpolate_wgan}, examples for the CelebA dataset are shown in Figures \ref{img:celebA_interpolate_wgan} and \ref{img:celebA_interpolate_wgan1}. In Figure \ref{img:celebA_interpolate128} we show bilinear interpolation of four random latent space samples, using the CelebA dataset with images of size 128$\times$128.

\subsection{Sample Diversity}
\begin{figure}[h]
	\centering
	\includegraphics[width=0.8\linewidth]{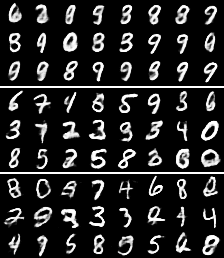}
	\caption{Random generated samples for the MNIST dataset. Latent vectors of the top three rows are sampled from $\mathcal{U}( -0.2, 0.2 )$, the three middle rows from $\mathcal{U}(-0.5, 0.5)$, and the bottom rows from $\mathcal{U}(-0.8, 0.8)$. The used networks are described in Table \ref{table:networks}, trained with WGAN-GP.}
	\label{img:mnist_wgan_uniform}
\end{figure}

\begin{figure*}[h]
	\centering
	\includegraphics[width=1\linewidth]{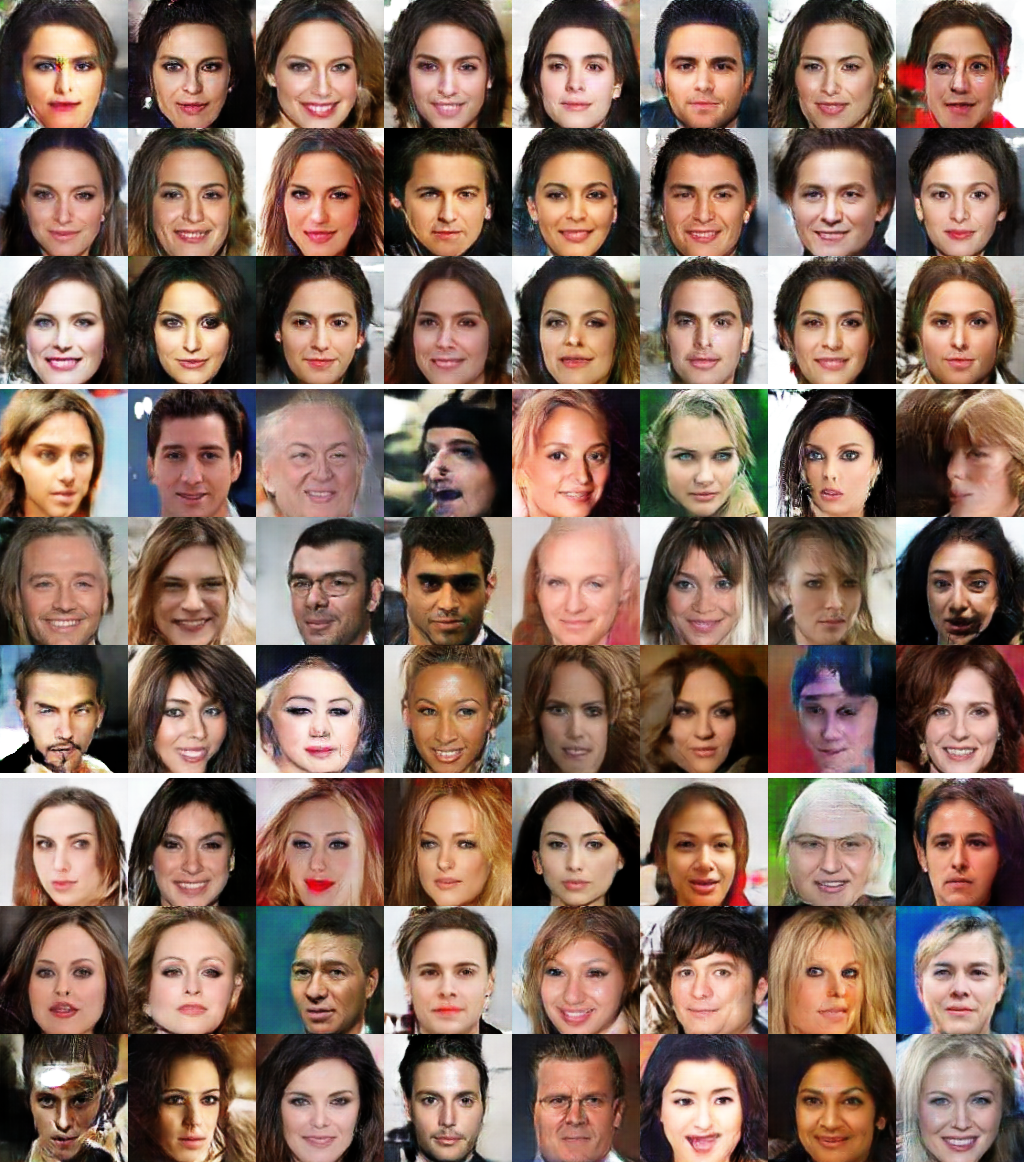}
	\caption{Random generated samples of size 128$\times$128 for the CelebA dataset. Latent vectors of top three rows are sampled from $\mathcal{U}( -0.2, 0.2 )$, the three middle rows from $\mathcal{U}(-0.5, 0.5)$, and the bottom rows from $\mathcal{U}(-0.8, 0.8)$. The used networks are described in Table \ref{table:networks}, trained with WGAN-GP.}
	\label{img:celebA_wgan_uniform}
\end{figure*}

During training we sample $\bm{z}$ from a uniform distribution $\mathcal{U}(-1,1)$. However, we found that the generated images depend on the sampled distribution. Figures \ref{img:mnist_wgan_uniform} and \ref{img:celebA_wgan_uniform} show the results of sampling the latent vector from different distributions (note that the networks are still trained on $\bm{z} \sim \mathcal{U}(-1, 1)$). We observe that when we sample from a distribution with a smaller range, the samples are of a higher quality but are less diverse. Considering $\mathcal{U}(-r, r)$, we can in general tradeoff diversity for quality. The higher $r$, the more diversity in the samples and the lower the quality.

\subsection{Conditional Attributes}
\begin{figure*}[h]
	\centering
	\includegraphics[width=1\linewidth]{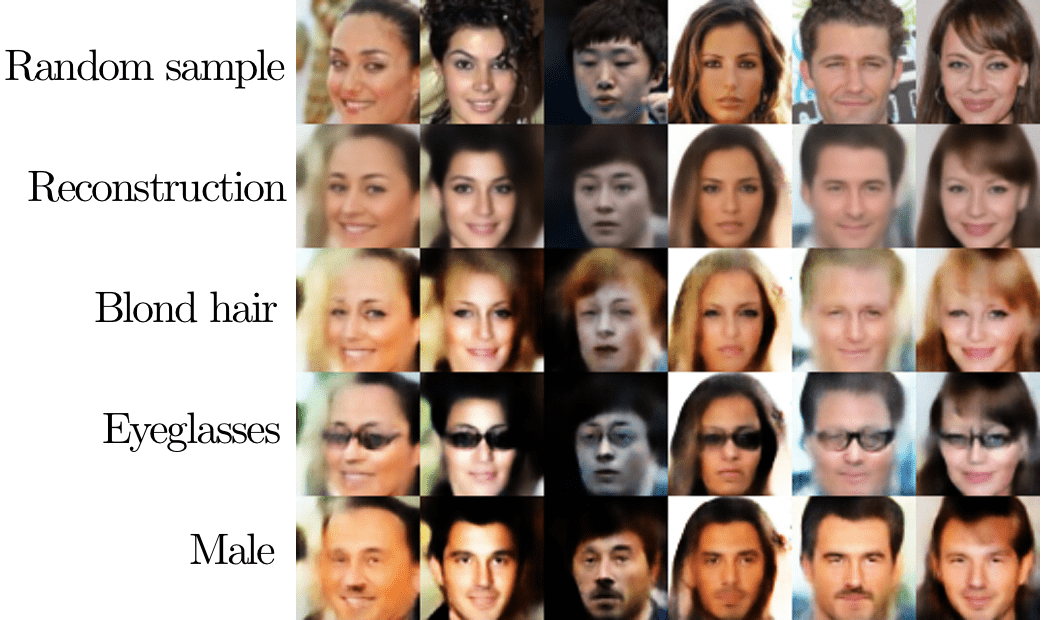}
	\caption{Random samples, encoded and conditioned on three different attributes. The used networks are described in Table \ref{table:networks}, trained with the conditional WGAN-GP.}
	\label{img:celebA_reconstruct_conditional}
\end{figure*}

\begin{figure*}[h]
	\centering
	\includegraphics[width=1\linewidth]{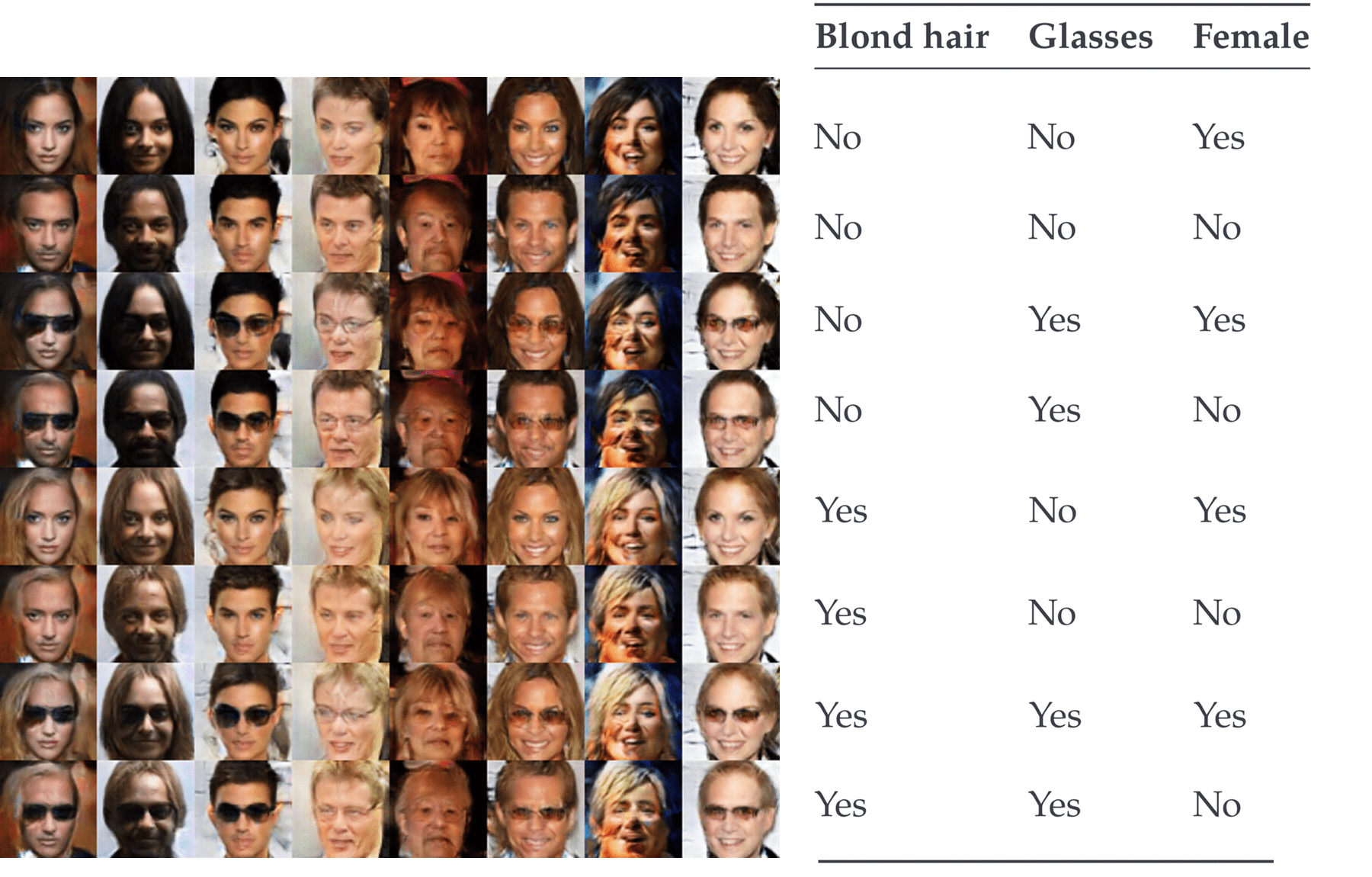}
	\caption{Random samples, conditioned on three different attributes showing all eight possible combinations. The used networks are described in Table \ref{table:networks}, trained with the conditional WGAN-GP.}
	\label{img:celebA_conditional_wgan}
\end{figure*}

By combining the encoder network with a conditional GAN, it is in theory possible to change visual attributes of a specific image. In Figure \ref{img:celebA_reconstruct_conditional} we demonstrate this. We start with encoding and reconstructing the original random samples. In order to change the visual attributes, we change the conditional variable $\bm{y}$ of the corresponding attribute. We feed this adapted conditional variable together with the encoded latent vector into the generator, producing the images in the bottom rows.\\
Figure \ref{img:celebA_conditional_wgan} shows that it is also possible to use multiple binary attributes at once. 

\end{appendices}

\end{document}